\pdfoutput=1

\documentclass[11pt]{article}

\usepackage[final]{acl}
\usepackage{graphics}
\usepackage{graphicx}
\usepackage{multirow}
\usepackage{booktabs}
\usepackage{makecell}
\usepackage{times}
\usepackage{latexsym}
\usepackage{graphics}
\usepackage{graphicx}
\usepackage{multirow}
\usepackage{amsthm,amsmath,amssymb}
\usepackage{mathrsfs}
\usepackage[T1]{fontenc}

\usepackage[utf8]{inputenc}

\usepackage{microtype}
\usepackage{url}
\title{A Simple yet Effective Relation Information Guided Approach \\ for Few-Shot Relation Extraction}

\author{
Yang Liu$^{\spadesuit\diamondsuit}$, Jinpeng Hu$^{\spadesuit\diamondsuit}$, Xiang Wan$^{\diamondsuit\clubsuit\dagger}$, Tsung-Hui Chang$^{\spadesuit\diamondsuit\dagger}$ \vspace{1.5mm} \\
\normalsize $^{\spadesuit}$The Chinese University of Hong Kong (Shenzhen)\\
\normalsize $^{\diamondsuit}$Shenzhen Research Institute of Big Data
\\
\normalsize $^{\clubsuit}$Pazhou Lab, Guangzhou, 510330, China \vspace{1mm} \\
\normalsize{
\texttt{
$^\spadesuit$\{yangliu5, jinpenghu\}@link.cuhk.edu.cn}}\\
\normalsize{
\texttt{$^{\diamondsuit}$wanxiang@sribd.cn \ $^{\spadesuit}$changtsunghui@cuhk.edu.cn}}
}

\begin{document}
\maketitle

\def\thefootnote{\dag}\footnotetext{Corresponding author.}
\renewcommand{\thefootnote}{\arabic{footnote}}

\begin{abstract}
%
Few-Shot Relation Extraction aims at predicting the relation for a pair of entities in a sentence by training with a few labelled examples in each relation. Some recent works have introduced relation information (i.e., relation labels or descriptions) to assist model learning based on Prototype Network. However, most of them constrain the prototypes of each relation class implicitly with relation information, generally through designing complex network structures, like generating hybrid features, combining with contrastive learning or attention networks. We argue that relation information can be introduced more explicitly and effectively into the model. Thus, this paper proposes a \textbf{direct addition} approach to introduce relation information. Specifically, for each relation class, the relation representation is first generated \textcolor{black}{by concatenating two views of relations (i.e., [CLS] token embedding and the mean value of embeddings of all tokens)} and then directly added to the original prototype for both train and prediction. Experimental results on the benchmark dataset FewRel 1.0 show significant improvements and \textcolor{black}{achieve comparable results to the state-of-the-art}, which demonstrates the effectiveness of our proposed approach. Besides, further analyses verify that \textbf{direct addition} is a much more effective way to integrate the relation representations and the original prototypes.
\footnote{The code is released at \url{https://github.com/lylylylylyly/SimpleFSRE}.}
\footnote{Main results in this paper can be found in the CodaLab competition (liuyang00) at \url{https://competitions.codalab.org/competitions/27980\#results}.}

\end{abstract}

\section{Introduction}

Relation Extraction (RE) \cite{bach2007review} is a fundamental task of Natural Language Processing (NLP), which aims to extract the relations between entities in sentences and can be applied to \textcolor{black}{other advanced tasks \cite{9359364, hu2021word}.}
%
However, RE usually suffers from labeling difficulties and train data scarcity due to the massive cost of labour and time. 
In order to solve the problem of data scarcity, Few-Shot Relation Extraction (FSRE) \cite{han2018fewrel, gao2019hybrid, qu2020few, yang2021entity} task has become a research hotspot in academia in recent years. The task is firstly to train on large-scale data on existing relation types, and then quickly migrate to a small amount of data on new relation types.
%
%

\begin{figure}
    \centering
    \includegraphics[width=0.48\textwidth, trim=0 10 0 5]{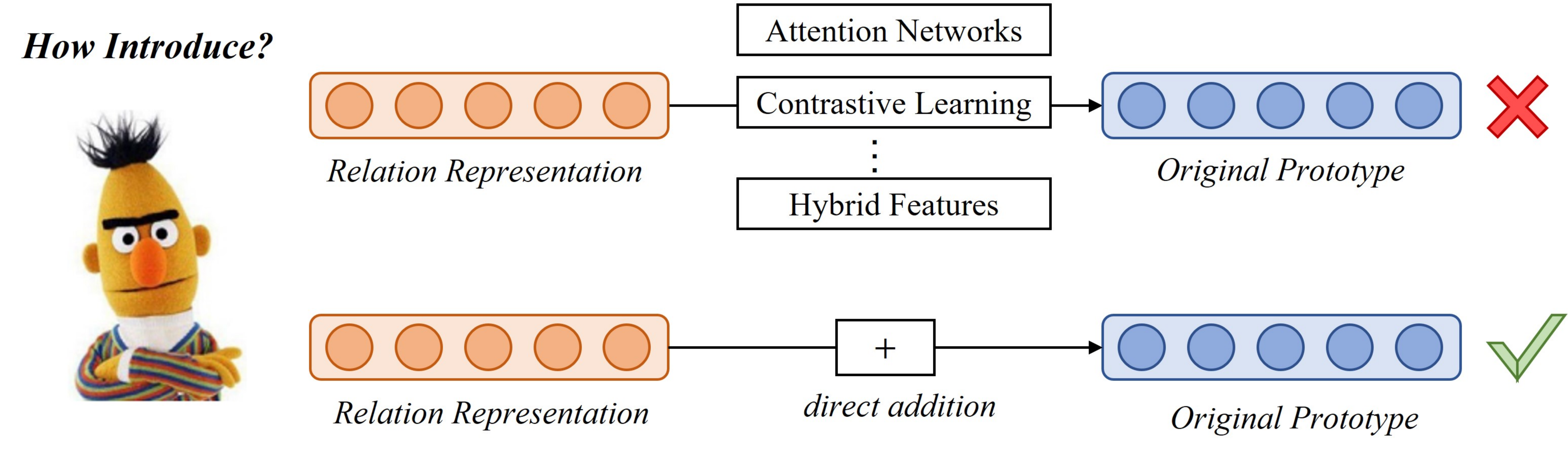}
    \caption{\textcolor{black}{The figure is an intuitive illustration of the difference in ways to introduce relation information between most existing works and our proposed approach. The orange vector and the blue vector denote representations of relations and prototypes, respectively. 
    }}
    \label{fig:1}
    \vskip -1.5em
\end{figure}

Inspired by the success of few-shot learning in
the computer vision (CV) community \cite{sung2018learning, garcia2018few}, various methods are introduced into FSRE. One of the popular algorithms is the Prototype Network \cite{snell2017prototypical}, which is based on the meta-learning framework \cite{vilalta2002perspective, vanschoren2018meta}. In detail, collections of few-shot tasks sampled from the external data containing disjoint relations are used as the training set for the model optimization.
For each few-shot task, the center of each relation class is calculated and used as the prototype of the relation class. Then, the model can be optimized by reducing the distances between the query samples and their corresponding prototypes.
Given a new sample, the model calculates which of the class prototypes is nearest to the new sample and assign it to this relation class.

%
In order to get better results, many works have utilized
relation information (i.e., relation labels or descriptions) to assist model learning. TD-proto \cite{yang2020enhance} enhanced prototypical network with both relation and entity descriptions. CTEG \cite{wang2020learning} proposed a model that learns to decouple high co-occurrence relations, where two types of external information are added.
%
Another intuitive idea is to hope that the model can learn good prototypes
or representations
, that is, to reduce the distances of the intra-class while widening the ones among different classes \cite{han2021exploring, dong2021mapre}, where  \citet{han2021exploring} introduced a novel approach
based on supervised contrastive learning that learns better prototype representations by the utilization of prototypes and relation labels and descriptions during the model training;  \citet{dong2021mapre} considered a semantic mapping framework, MapRE, which leverages both label-agnostic and label-aware knowledge in pre-training
and fine-tuning processes.

%
\textcolor{black}{However, there are two limitations in how these works introduce relation information. The first is that most of them take implicit constraints, like contrastive learning or relation graphs, instead of the direct fusion, which can be weak facing the remote samples. The second is that they usually adopt complicated designs or networks, like hybrid features or elaborate attention networks, which can bring too many or even harmful parameters.
%
%
Therefore, in this paper, we propose a straightforward yet effective way to incorporate the relation information into the model.
}
%
%
\textcolor{black}{Specifically, on the one hand, the same encoder is used to encode relation information and sentences for mapping them into the same semantic space.
On the other hand, we generate the relation representation for each relation class by concatenating two relation views (i.e., [CLS] token embedding and the mean value of embeddings of all tokens), which allows relation representations and prototypes to form the same dimension.
Afterwards, the generated relation representation is directly added to the prototype for enhancing model train and prediction. %
}

Figure \ref{fig:1} shows an intuitive illustration of the difference in ways to introduce relation information between most existing works and our proposed approach. Based on the mentioned two limitations of previous works, we provide two possible high-level ideas about why our proposed approach should work for few-shot relation extraction. The first is that the  \textbf{direct addition} is a more robust way to generate promising prototypes than implicit constraints when facing the remote samples. The second is that the \textbf{direct addition} does not bring extra parameters and simplifies the model. Due to possible over-fitting, fewer parameters are always better than more parameters, especially for few-shot tasks. We conduct experimental analyses in the experiment section for further demonstration.
%

We conduct experiments on the popular FSRE benchmark FewRel 1.0 \cite{han2018FewRel} under four few-shot settings. Experimental results show considerable improvements and \textcolor{black}{achieve comparable results to the state-of-the-art}, which demonstrates the effectiveness of our proposed approach, i.e., the direct addition operation.




\begin{figure}
    \centering
    \includegraphics[width=0.45\textwidth, trim=0 6 0 5]{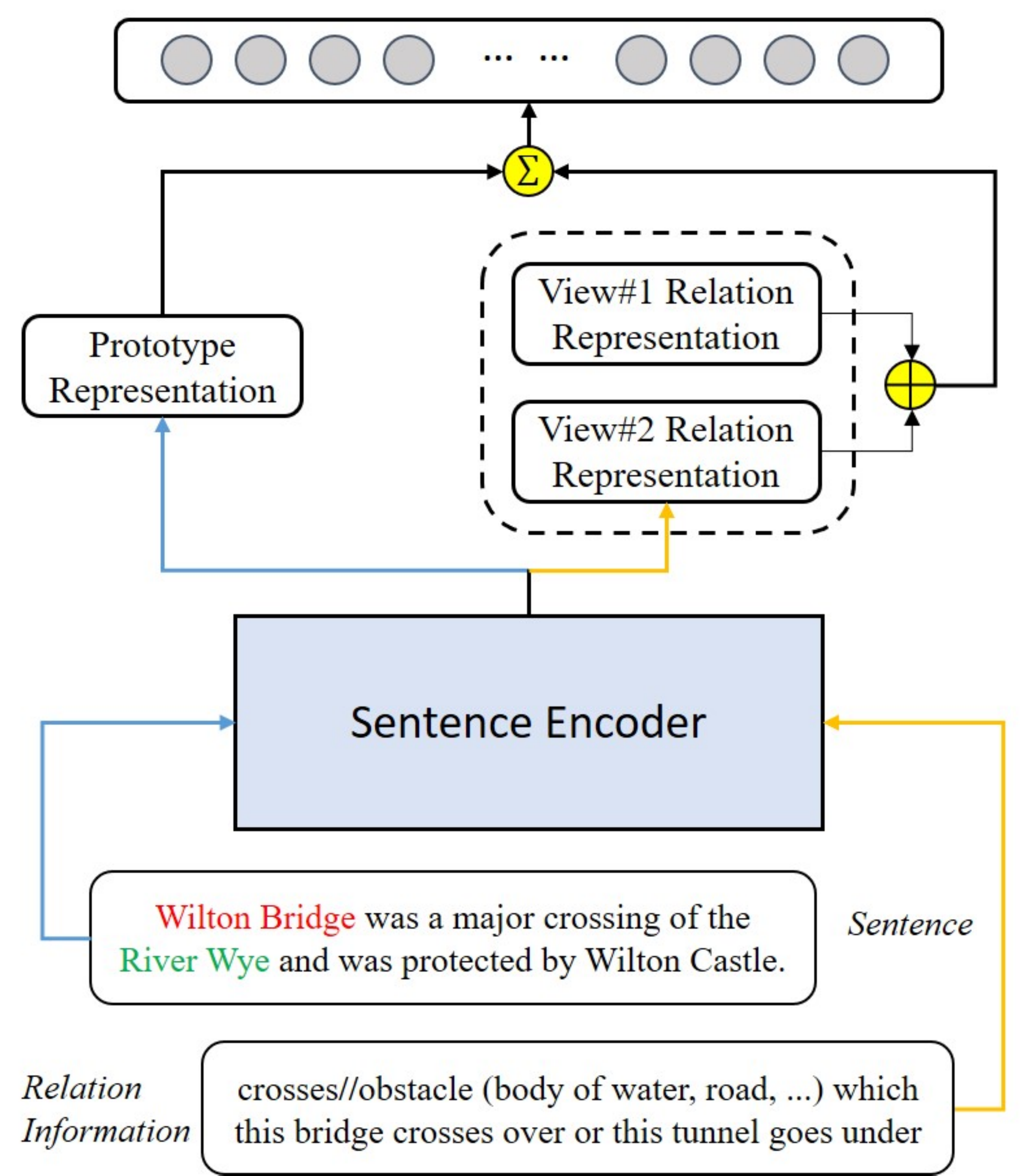}
    \caption{The overall structure of our proposed approach, in which the sentence and the relation information share the same encoder, and then the relation representation is generated through $\bigoplus$ operation with two views of relations and added to the original prototype representation. $\bigoplus$ and $\sum$ denote the concatenation and addition operations, respectively.}
    \label{fig:structure}
    \vskip -0.5em
\end{figure}


\begin{table*}[!]
\centering
\resizebox{.88\textwidth}{!}{
\begin{tabular}{llrrrr}
\hline
\textbf{Encoder} & \textbf{Model} & \textbf{5-w-1-s} & \textbf{5-w-5-s} & \textbf{10-w-1-s} & \textbf{10-w-5-s} \\
\hline
\multirow{2}*{CNN} & Proto-HATT & 72.65 / 74.52 & 86.15 / 88.40 & 60.13 / 62.38 & 76.20 / 80.45 \\
& MLMAN & 75.01 / --- --- & 87.09 / 90.12 & 62.48 / --- --- & 77.50 / 83.05 \\ \hline
\multirow{12}{*}{BERT} &  
BERT-PAIR & 85.66 / 88.32 & 89.48 / 93.22 & 76.84 / 80.63 & 81.76 / 87.02\\
 & Proto-BERT$^*$ & 84.77 / 89.33 & 89.54 / 94.13 & 76.85 / 83.41 & 83.42 / 90.25 \\
& REGRAB & 87.95 / 90.30 & 92.54 / 94.25 &80.26 / 84.09 & 86.72 / 89.93 \\
& TD-proto &--- --- / 84.76 & --- --- / 92.38 & --- --- / 74.32 & --- --- / 85.92 \\
& CTEG & 84.72 / 88.11 & 92.52 / 95.25 & 76.01 / 81.29 & 84.89 / 91.33 \\
& ConceptFERE &--- --- / 89.21 & --- --- / 90.34 & --- --- / 75.72 & --- --- / 81.82 \\
& HCRP (BERT) & 90.90 / 93.76 & 93.22 / 95.66 & 84.11 / 89.95 & 87.79 / 92.10 \\
& Ours (BERT) & 91.29 / \textbf{94.42} & 94.05 / \textbf{96.37} & 86.09 / \textbf{90.73} & 89.68 / \textbf{93.47} \\\cline{2-6}
& MTB &--- --- / 91.10 & --- --- / 95.40 & --- --- / 84.30 & --- --- / 91.80 \\
& CP  &--- --- / 95.10 & --- --- / 97.10 & --- --- / 91.20 & --- --- / 94.70 \\
& MapRE &--- --- / 95.73 & --- --- / 97.84 & --- --- / 93.18 & --- --- / 95.64 \\ 
& HCRP (CP) & 94.10 / 96.42 & 96.05 / \textbf{97.96} & 89.13 / 93.97 & 93.10 / \textbf{96.46} \\
&Ours (CP) & 96.21 / \textbf{96.63} & 97.07 / 97.93 & 93.38 / \textbf{94.94} & 95.11 / 96.39 \\ \hline
 & $\Delta$ &  +5.09 & +2.24 & +7.32 & +3.22 \\
& $\Delta$ (CP) & +1.53 & +0.83 & +3.74 & +1.69 \\
\hline
\end{tabular}}
\caption{\label{main-results}
Experimental results of FSRE on FewRel 1.0 validation / test set, where $N$-w-$K$-s stands for the abbreviation of $N$-way-$K$-shot. The table divides the method with BERT as the encoder into two parts, from top to bottom including approaches with the original BERT, and approaches with additional pre-training on BERT. Note that $*$ represents the results of our implementation, others are obtained from results reported by papers or CodaLab.}
\end{table*}

\section{Proposed Method}
In this section, we present the details of our proposed approach. Figure \ref{fig:structure} shows the overall structure, where the blue and yellow lines represent the flow of sentences and relation information, respectively. In order to map the representations of sentences and relation information into the same semantic space, the shared sentence encoder is utilized. 
Then, we concatenate two views of the relation representations for obtaining the same dimension as prototypes and integrate relation representations into original prototypes by direct addition.


\subsection{Sentence Encoder}
\label{2.1}
We employ one BERT \cite{devlin-etal-2019-bert} as the encoder to obtain contextualized embeddings of support set $\mathcal{S}$ and query set $\mathcal{Q}$.
For instances in $\mathcal{S}$ and
$\mathcal{Q}$, 
\textcolor{black}{\textcolor{black}{the intermediate states} are obtained by concatenating the hidden states corresponding to start tokens
of two entity mentions following \citet{baldini-soares-etal-2019-matching}}, i.e., $[h_{entity1};h_{entity2}]$, $h_{entity1}, h_{entity2} \in \mathbb{R}^d$, where $d$ is the size of the contextualized representations of sentence encoder. Then, we average intermediate states of each relation class in $\mathcal{S}$ to obtain the initial prototype representation for each relation class. Denote the set of prototype representations as $\{\mathcal{P}_i\in\mathbb{R}^{2d};i=1,2,..,c\}$, where $c$ is the number of relation classes. 
%
%
For each relation, we concatenate the name and description and feed the sequence into the BERT encoder.
We treat the embedding of the "[CLS]" token, i.e., $\{\mathcal{R}^{view1}_i\in\mathbb{R}^d, i=1,2,...,c\}$, and the average of the embeddings of all tokens, i.e., $\{\mathcal{R}^{view2}_i\in\mathbb{R}^d, i=1,2,...,c\}$, as two different views from the relation representation.

\vspace{-1mm}
\subsection{Relation Representation Generation}
\vspace{-0.5mm}
\label{ourmethod2.2}
As described in Section \ref{2.1}, $\mathcal{P}_i\in\mathbb{R}^{2d}$ for prototypes and $\mathcal{R}^{view1}_i, \mathcal{R}^{view2}_i\in\mathbb{R}^d$ for relations. In order to minimize the introduction of additional linear layers (or parameters) and make the direct addition operation possible, we combine $\mathcal{R}^{view1}$ and $\mathcal{R}^{view2}$ together by simple concatenation operation $\oplus$ as the following.


\begin{equation}
    \mathcal{R}^{final} = \mathcal{R}^{view1} \oplus \mathcal{R}^{view2}
\end{equation}
where $\mathcal{R}^{final}\in \mathbb{R}^{2d}$ same as $\mathcal{P}$.

\subsection{Relation Classification}
The final prototype representations are obtained by the direct addition of the original prototype representations $\mathcal{P}$ and the relation representations $\mathcal{R}^{final}$:
\begin{equation}
    \mathcal{P}^{final} = \mathcal{P} + \mathcal{R}^{final} = \{\mathcal{P}_i^f\in \mathbb{R}^{2d} \}
\end{equation}
The model uses the \textbf{vector dot product} way to calculate the distance between the query instance $\mathcal{Q}$ and each class prototype $\{\mathcal{P}_i^{final} \in \mathbb{R}^{2d}, i=1,2,..,c\}$, and selects the relation class with the shortest distance as the prediction result. We employ the cross-entropy (CE) loss as the loss function simply:
\begin{equation}
    \mathcal{L}_{CE} = -log(z_y)
\end{equation}
where $y$ is the class label, and $z_y$ is the estimated probability for the class $y$.

\section{Experiment}

\subsection{Dataset, Training, Evaluation and Comparable Models}
\paragraph{Dataset} Our proposed approach is evaluated on the commonly used large-scale FSRE dataset FewRel 1.0 \cite{han2018FewRel}, which consists of 100 relations, each with 700 labeled instances. Our experiments follow the splits used in official benchmarks, which split the dataset into 64 base classes for training, 16 classes for validation, and 20 novel classes for testing. 

\vspace{-1mm}

\paragraph{Training} We use BERT-base-uncased and CP \cite{wang2020learning} as the sentence encoder, where CP is a further pre-trained model based on BERT with contrastive learning. We set the train iteration number as 30,000, validation iteration number as 1,000, batch size as 4, learning rate as 1e-5 and 5e-6 for BERT and CP respectively.
\vspace{-1mm}

\begin{table}[!]
\centering
\resizebox{.40\textwidth}{!}{
\begin{tabular}{lccc}
\hline
Settings & CP & HCRP (CP) & Ours (CP) \\ \hline
5-w-1-s & 95.10 & 96.42 & 96.63\textcolor{red}{$\uparrow$}\\ 
5-w-5-s & 97.10 & 97.96 &  96.93\textcolor{green}{$\downarrow$}\\
10-w-1-s & 91.20 & 93.97 & 94.94\textcolor{red}{$\uparrow$}\\
10-w-5-s & 94.70 & 96.46 & 96.39\textcolor{green}{$\downarrow$} \\ \hline
Average & 94.53 & 96.20 & 96.47\textcolor{red}{$\uparrow$} \\
\hline
\end{tabular}}
\caption{\label{HCRP} The comparison with HCRP on the test set.
}
\end{table}

\begin{table}[!]
\centering
\resizebox{.33\textwidth}{!}{
\begin{tabular}{ccc}
\hline
  &HCRP & Ours \\ \hline
Para. & 110.66M & 109.48M  \\ \hline
\multicolumn{3}{c}{Parameters to be adjusted} \\ \hline
\multirow{3}{*}{Training} &   \multicolumn{2}{c}{learning rate}  \\
 &   \multicolumn{2}{c}{batch size}  \\
  &   \multicolumn{2}{c}{max iteration} \\ \hline
\multirow{2}{*}{Loss} & $\lambda$ & \multirow{2}{*}{none} \\
& $\gamma$ &  \\
\hline
\end{tabular}}
\caption{\label{para} Comparison on the model complexity.
}
\end{table}

\paragraph{Evaluation} $N$-way-$K$-shot ($N$-w-$K$-s) is commonly used to simulate the distribution of FewRel in different situations, where $N$ and $K$ denote the number of classes and samples from each class, respectively. In the $N$-w-$K$-s scenario, accuracy is used as the performance metric. Since the label of the test set of the FewRel is not publicly available, we submit the prediction of our model to \textcolor{black}{CodaLab} to obtain the accuracy on the test set.

\vspace{-1mm}

\paragraph{Comparable Models} The comparable models contain two CNN-based models Proto-HATT \cite{gao2019hybrid} and MLMAN \cite{ye2019multi}, as well as nine BERT-based models BERT-PAIR \cite{gao2019FewRel}, REGRAB \cite{qu2020few}, TD-proto \cite{yang2020enhance}, CTEG \cite{wang2020learning}, ConceptFERE \cite{yang2021entity}, MTB \cite{baldini-soares-etal-2019-matching}, CP \cite{peng2020learning}, MapRE \cite{dong2021mapre}, and HCRP \cite{han2021exploring}. Since our proposed approach is based on the Prototype Network with BERT, we also compare the Proto-BERT without relation information.

\subsection{Results}

All experimental results are shown in Table \ref{main-results}. CNN-based and BERT-based methods are both contained in the table. There are two parts to BERT-based methods. 
The first one utilizes the original BERT without any external pre-training. \textit{Proto-BERT} represents the method on which our model is based, which means that this is the result of the model without introducing the improvements we propose. This result will also be analyzed and displayed again in Section \ref{ablation}.  %
The second one contains the methods that employ additional pre-training on BERT with Wikipedia data or contrastive learning to get better contextual representations. 
We apply our approach to BERT and CP. For obvious comparison, the former is shown in the first part of BERT-based models, and the latter is shown in the second part of BERT-based models. The \textbf{last two rows} show the increase on the test set compared to the basic models used by our approach (i.e., Proto-BERT and CP).
%
%

From the table, we can obtain three observations. 
First, when using BERT as the backend model, our approach \textit{Ours (BERT)} outperforms the state-of-the-art, which is listed in the first part of the BERT-based model in Table \ref{main-results}. Most of these methods are designed with relatively complex network structures and implementations.
Second, \textit{Ours (CP)} utilizes CP as the backend model and outperforms the state-of-the-art, i.e., HCRP (CP), on two few-shot settings, i.e., 5-way-1-shot and 10-way-1-shot, which also reflects from the side that our approach is more suitable for few-shot scenarios.
%
Third, the improvements compared to the basic model, i.e., Proto-BERT and CP, are rather considerable, which are shown in the last two rows of Table \ref{main-results}.
These observations demonstrate the effectiveness of our proposed approach.
\paragraph{Comparision with HCRP}

In this part, we compare our approach with the state-of-the-art model, i.e., HCRP, on the test set based on CP, which is shown in Table \ref{HCRP}. It can be seen that the result of our approach is slightly lower than HCRP on 5-way-5-shot and 10-way-5-shot, while the average accuracy of four settings is higher than HCRP.
However, HCRP designed three modules, including hybrid features, contrastive learning, and task adaptive loss function.
On the contrary, our approach is more straightforward and achieves comparable results to HCRP. Table \ref{para} shows the comparison of the model complexity between Ours and HCRP. It can be seen that the total number of model parameters and parameters to be adjusted of Ours are both less than HCRP.  
Thus, we may argue that the lower results on 5-way-5-shot and 10-way-5-shot than HCRP can not deny the effectiveness of our proposed approach.

\begin{table}[!]
\centering
\begin{tabular}{lrr}
\hline
Model & 5-w-1-s  & 10-w-1-s \\ \hline
Ours & 91.29  & 86.09 \\ \hline
w/o relation info.& 84.77 & 76.85 \\ \hline
w/ concat & 79.16 & 65.12 \\
w/ linear layer &  &  \\
\quad \quad view$\#$1 & 89.04 & 80.29 \\
\quad \quad view$\#$2 & 89.39 & 80.14\\
\hline
\end{tabular}
\caption{\label{ablationresult} Ablation Study on validation set of FewRel 1.0. w/o, w/, and info. are the abbreviations of without, with, and information.
}
\vskip -1em
\end{table}


\subsection{Ablation Study}
\label{ablation}
Since the label of the test set of FewRel 1.0 is not public, in this section, we conduct an ablation study on 5-way-1-shot (5-w-1-s) and 10-way-1-shot (10-w-1-s) based on BERT with the validation set. Following HCRP and the official setting, the validation iteration step is set to 1000. Results are shown in Table \ref{ablationresult}.
There are two types of ablation study.  
One type is "w/o relation info.", where only the original prototype network is utilized without introducing any relation information (i.e., Proto-BERT).
The second type is ablation study in the integration way of relation representations and prototypes. Instead of the direct addition operation, we adopt another two kinds of integration way, i.e., "w/ concat" and "w/ linear layer". For “w/ concat”, after obtaining the relation representation $\mathcal{R}^{final} \in \mathbb{R}^{2d}$ (the symbol appeared in Section \ref{ourmethod2.2}) with two views of relations, we perform "w/ concat" by concatenating $\mathcal{R}^{final}$ and $\mathcal{P}$ first, i.e., $\mathcal{R}^{final}\oplus\mathcal{P} \in \mathbb{R}^{4d}$. Then a $4d$-$2d$ linear layer is applied on the concatenation embedding to obtain the final prototype representation. For "w/ linear layer", a extra linear layer is introduced. Specifically, only one view of relations, i.e.,  $\mathcal{R}^{view1}\in\mathbb{R}^d$ or $\mathcal{R}^{view2}\in\mathbb{R}^d$, is used in the model. Then, we perform a $1d$-$2d$ linear layer and addition operation to obtain the final prototype representation.
%
%

From the results in Table \ref{ablationresult}, we can see that relation information is essential for few-shot relation extraction. The result ("w/o relation info.") drops sharply compared to "Ours". Besides, results of another integration way have poor performance compared to "Ours". 
Especially, "w/ concat" are quite poor, probably because it requires the use of a $4d$-$2d$ linear layer, which introduces too many parameters. These observations demonstrate that our proposed approach is a straightforward yet effective way to integrate relation representations and original prototypes. 


\section{Conclusion}
\vspace{-1mm}
In this paper, we proposed a simple yet effective approach with relation information based on Prototype Network. The core idea is to introduce relation representations by the direct addition operation instead of designing complex structures.
Experimental results on FewRel 1.0 achieve comparable results to the state-of-the-art and demonstrate the effectiveness of our proposed approach. Moreover, we provide two high-level ideas, i.e., explicit constraints and fewer parameters, about why the direct addition is so effective. We believe that the idea of finding global information to perform the direct addition with original prototypes is general and can be extended to other few-shot tasks that can be modeled based on the prototype network.

Since the direct addition way of introducing relations is simple and efficient, we also believe that future work should focus more on generating better relation representations rather than designing fusion methods between relations and prototypes.
\section*{Acknowledgments}
\textcolor{black}{This work is supported by Chinese Key-Area Research and Development Program of Guangdong Province (2020B0101350001), NSFC under the project “The Essential Algorithms and Technologies for Standardized Analytics of Clinical Texts” (12026610) and the Guangdong Provincial Key Laboratory of Big Data Computing, The Chinese University of Hong Kong, Shenzhen.}
\newpage
\bibliography{anthology}
\bibliographystyle{acl_natbib}

\newpage
\section*{Appendix}
\subsection*{A. Hyper-parameter Settings}
Hyper-parameter settings for two backend models, i.e., BERT and CP, are shown in Table \ref{Tab:BERT-hyperparameters} and Table \ref{Tab:CP-hyperparameters}.
We experiments the model on three different learning rate and select the best learning rate that is bolded in the table. Besides, the validation iteration is set to 1000. 

\begin{table}[h]
\small
\centering
\resizebox{.49\textwidth}{!}{
\begin{tabular}{l|l|c}
\toprule[1pt]
{\textsc{\textbf{Component}}}&\textsc{\textbf{Parameter}}  &\textsc{\textbf{Value}} \\
\hline                       
\multirow{3}{*} {\makecell*[l]{\textsc{BERT}}}
& \textsc{type} & {base-uncased} \\
& \textsc{hidden size} & 768  \\
& \textsc{max length} & 128  \\
\midrule
\multirow{3}{*} {\makecell*[l]{\textsc{Training}}}
& \textsc{learning rate} & 9e-6,\textbf{1e-5}, 2e-5 \\
& \textsc{batch size} & 4  \\
& \textsc{max iterations} & 30,000  \\
\bottomrule
 \end{tabular}}
  \caption{The hyper-parameters that we have experimented on the datasets with BERT.}%
  \label{Tab:BERT-hyperparameters}
\end{table}

\begin{table}[h]
\small
\centering
\resizebox{.49\textwidth}{!}{
\begin{tabular}{l|l|c}
\toprule[1pt]
{\textsc{\textbf{Component}}}&\textsc{\textbf{Parameter}}  &\textsc{\textbf{Value}} \\
\hline                       
\multirow{3}{*} {\makecell*[l]{\textsc{CP}}}
& \textsc{type} & {base-uncased} \\
& \textsc{hidden size} & 768  \\
& \textsc{max length} & 128  \\
\midrule
\multirow{3}{*} {\makecell*[l]{\textsc{Training}}}
& \textsc{learning rate} & \textbf{5e-6}, 7e-6, 9e-6 \\
& \textsc{batch size} & 4  \\
& \textsc{max iterations} & 30,000  \\
\bottomrule
 \end{tabular}}
  \caption{The hyper-parameters that we have experimented on the datasets with CP.}%
  \label{Tab:CP-hyperparameters}
\end{table}

\subsection*{B. Results on Different Learning Rates}
We explore the effect of two different learning rates with CP as the backend model, which is shown in Table \ref{differ-lr}. Note that \textit{lr} is short for learning rate. From the results in Table \ref{differ-lr}, we can see that when CP is used as the backend model, our method has better performance with a smaller learning rate.
\begin{table}[h]
\centering
\resizebox{.35\textwidth}{!}{
\begin{tabular}{lcc}
\hline
Settings & \textit{lr}=5e-6 & \textit{lr}=9e-6  \\ \hline
5-w-1-s & 96.63 & 96.54 \\ 
5-w-5-s & 97.93 & 97.98\\
10-w-1-s & 94.94 & 94.04 \\
10-w-5-s & 96.39 & 96.08 \\ \hline
Average & 96.47 & 96.16\\
\hline
\end{tabular}}
\caption{\label{differ-lr} Test accuracy on four settings with two different learning rates based on CP.
}
\end{table}

\subsection*{C. Comparison with different modules of HCRP}
HCRP contains three modules including hybrid features generation, relation-prototype contrastive learning (RPCL),  and task adaptive loss function. HCRP reported the ablation study in the paper with different modules. We further do the comparison with different modules of HCRP on FewRel 1.0 validation set based on BERT model. The comparisons are based on 5-way 1-shot and 10-way 1-shot settings.
\begin{table}[h]
\small
\centering
\begin{tabular}{l|rrr}
\toprule[1pt]
{{\textsc{\textbf{Model}}}}&\textsc{\textbf{5-w-1-s}}  &\textsc{\textbf{10-w-1-s}} \\
\hline                       
 {\makecell*[l]{Ours}}
& 91.29 & 86.09 \\ \hline
{{\textsc{\textbf{HCRP}}}} & 90.90 & 84.11 \\
w/o local prototype & 88.37 & 82.31 \\
w/o global prototype & 86.42 & 77.86 \\
w/o RPCL & 87.85 & 79.76 \\
w/o task adaptive loss & 88.96 & 82.75 \\
\bottomrule
 \end{tabular}
  \caption{Comparisons with different modules of HCRP on FewRel 1.0 validation set with BERT model, where w/o denotes without.}%
  \label{Tab:hyperparameters}
\end{table}


\end{document}